\documentclass[11pt,a4paper]{article}
\usepackage[hyperref]{acl2020}
\usepackage{times}
\usepackage{latexsym}
\usepackage{algorithm2e}
\usepackage{amsfonts, amsmath}
\usepackage{amsthm, graphicx}
\usepackage{amssymb, makecell}
\usepackage{kotex}

\usepackage{microtype}

\aclfinalcopy 
\def\aclpaperid{732} 

\title{Graph Sequential Network for Reasoning over Sequences}

\author{Ming Tu, Jing Huang, Xiaodong He, Bowen Zhou \\
  JD AI Research \\
  \texttt{\{ming.tu,jing.huang,xiaodong.he,bowen.zhou\}@jd.com} \\}

\date{}

\begin{document}
\maketitle
\begin{abstract}
Recently Graph Neural Network (GNN) has been applied successfully to various NLP tasks that require reasoning, such as multi-hop machine reading comprehension. 
In this paper, we consider a novel case where reasoning is needed over graphs built from sequences, i.e. graph nodes with sequence data. Existing GNN models fulfill this goal by first summarizing the node sequences into fixed-dimensional vectors, then applying GNN on these vectors. To avoid information loss inherent in the early summarization and make sequential labeling tasks on GNN output feasible, we propose a new type of GNN called Graph Sequential Network (GSN), which features a new message passing algorithm based on co-attention between a node and each of its neighbors. We validate the proposed GSN on two NLP tasks: interpretable multi-hop reading comprehension on HotpotQA and graph based fact verification on FEVER. Both tasks require reasoning over multiple documents or sentences. Our experimental results show that the proposed GSN attains better performance than the standard GNN based methods.
\end{abstract}

\section{Introduction}

Graph neural network (GNN) has attracted much attention recently,
and have been applied to various tasks such as bio-medicine \cite{zitnik2018modeling}, computational chemistry \cite{gilmer2017neural}, social networks \cite{fan2019graph}, computer vision \cite{li2018beyond}, and natural language understanding \cite{xiao2019dfgn, tu2019hdegraph}. GNN assumes structured graphical inputs, for example, molecule graphs, protein-protein interaction networks, or language syntax trees, which can be represented with a graph $\mathcal{G} = (\mathcal{V}, \mathcal{E})$. $\mathcal{V}$ defines a set of nodes and $\mathcal{E}$ defines a set of edges, each of which connecting two different nodes in $\mathcal{V}$. 

Different GNN variants have been proposed to learn graph representation,
which include Graph Convolutional Network (GCN) \cite{kipf2016semi}, GraphSage \cite{hamilton2017inductive}, Graph Isomorphism Network (GIN) \cite{xu2018powerful} and Graph Attention Network (GAT) \cite{velivckovic2017graph}. Existing GNN variants assume features of each node to be a vector, which is initialized by predefined features or learnt by feature encoding networks. In cases where each node $v$ is represented by a sequence of feature vectors, usually in natural language processing (NLP) tasks, common practice would take the encoded sequential feature vectors, and go through a summarization module that is either based on simple average/max pooling or parametric attentive pooling to convert the sequential feature vectors to a fixed-dimensional feature vector. Then GNN-based message passing algorithm is applied to obtain node representations from these summarized feature vectors \cite{tu2019hdegraph,xiao2019dfgn,zhou2019gear}.

However, this early summarization strategy (summarization before GNN based representation learning) could bring inevitable information loss \cite{seo2016bidirectional}, and result in 
information flow bottleneck thus less powerful reasoning ability among graph nodes. Furthermore, early summarization also makes sequential labeling tasks impossible because GNN only outputs one vector for each input sequence, while sequential labeling tasks need sequential inputs.

\begin{figure*}[t]
    \centering
    \includegraphics[width=0.9\linewidth]{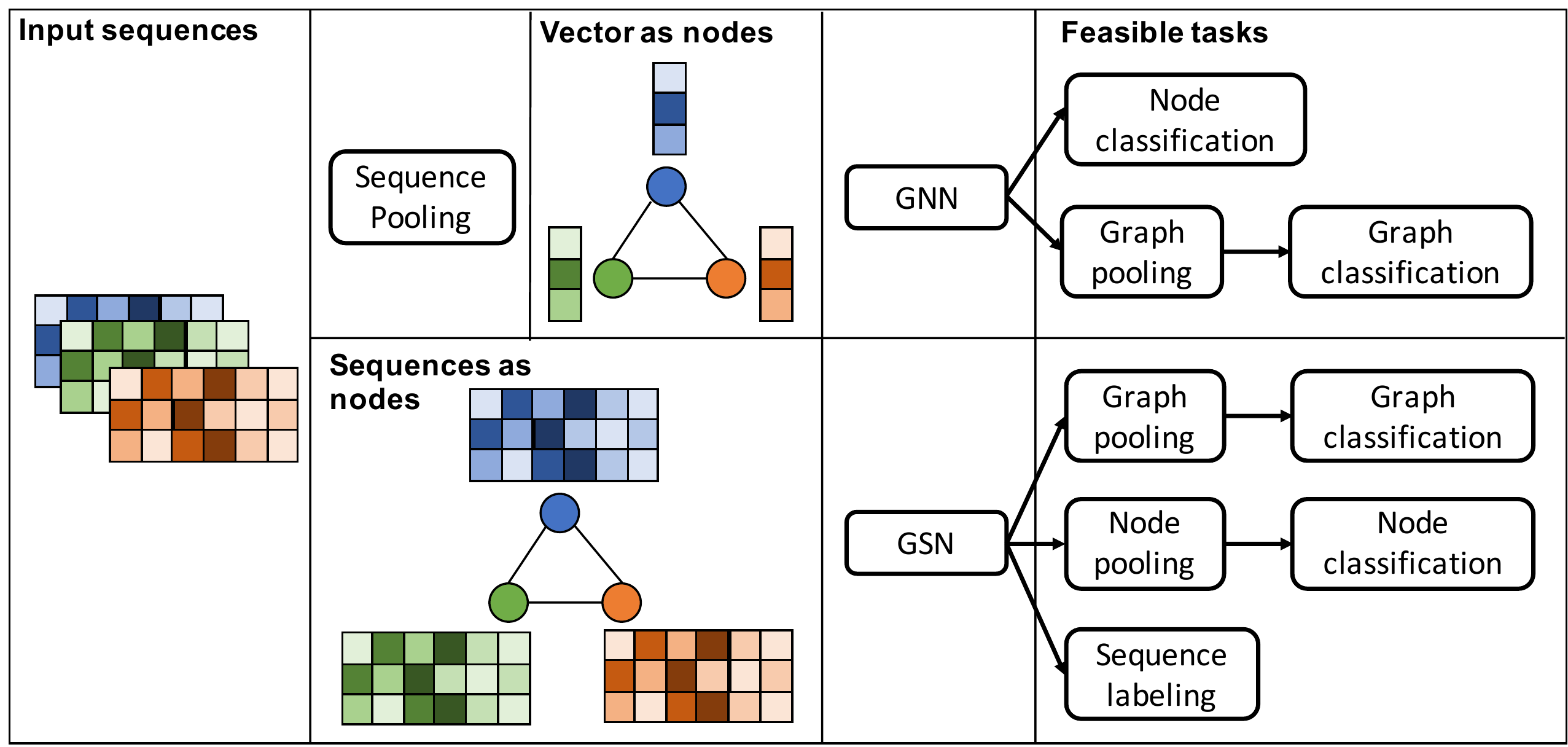}
    \caption{Diagram of the proposed GSN and its comparison with GNN when dealing with a graph built from multiple sequences (3 as in the figure). With the same input, the first row demonstrates the common pipeline of GNN based models, and the second row is the pipeline of our proposed GSN based models. We also show feasible tasks, and our proposed GSN can tackle sequential labeling tasks while GNN can not.}
    \label{fig:diagram}
\end{figure*}

To alleviate these limitations, in this paper we propose a new type of GNN: Graph Sequential Network (GSN) to directly learn feature representations over graphs with a sequence for each node. GSN differs from previous GNN variants in the following way:
\begin{enumerate}
    \item GSN can directly conduct message passing over nodes represented with sequential feature vectors, thus avoid information loss due to the pooling for early summarization.
    \item Both the input and output of the proposed GSN are sequences, making sequential labeling tasks on GSN output possible.
\end{enumerate}
  To achieve these advantages, we propose a new message passing algorithm based on co-attention between a node and each of its neighbors. Co-attention is commonly used in NLP tasks, especially in machine reading comprehension (MRC), as a way to encode query-aware contextual information based on affinity matrix between two sequences \cite{xiong2016dynamic, seo2016bidirectional, zhong2019coarse}. In the context of this paper, the advantage of co-attention is that it can encode neighbor-aware information of the current node represented by a sequence of feature vectors, even when neighbors have different sequence lengths. The learned sequential representation of each node can then be used for node-level sequence classification or sequential labeling, or graph-level classification tasks. The general idea of our proposed GSN and its comparison with existing GNN based methods is shown in Figure~\ref{fig:diagram}.

To validate the effectiveness of the proposed GSN, we experiment on two NLP datasets: HotpotQA \cite{yang2018hotpotqa} and fact extraction and verification data set provided by FEVER shared task 1.0 \cite{thorne2018fever}. Both tasks require the model to have reasoning ability, and top performance has been achieved with early summarization followed by GNN \cite{zhou2019gear, xiao2019dfgn, fang2019hierarchical, tu2019select}. With thorough experiments, we show that the proposed GSN achieves better performance than standard GNN, proving its stronger ability to do reasoning over sequences.

\section{Related Work}
GNN has been proposed as powerful models for graph representation learning. Different from Convolutional Neural Networks (CNN) which work on Euclidean space, GNN operates on graph data, which are usually defined as a set of graph nodes and the edges connecting those nodes. GNN implements neural-network-like message passing algorithms to update graph node representation from each node's neighborhood. The resulting node representations encode structural information from the subgraph within $k$ hops away from each node. 

Multiple GNN variants have been proposed with different message passing algorithms. For example GCN \cite{kipf2016semi}, Graph Sage\cite{hamilton2017inductive}, GAT\cite{velivckovic2017graph}, GIN\cite{xu2018powerful}, etc. Our proposed GSN can be regarded as a variant of GNN. However, GSN differs from previous GNN variants in that GSN operates on graphs with sequences as nodes. Thus GSN needs a new message passing algorithm for nodes represented with sequences.

\noindent\textbf{GNN for NLP:} recently various research work on NLP adopted GNN and gained benefit from them.
These work can be roughly categorized into two groups depending on the way to build graphs. 
The first group usually builds graphs from parsing trees or develops graph-like Recurrent Neural Networks (RNN). \citet{bastings2017graph} and \citet{marcheggiani2018exploiting} explored building graphs from syntactic or semantic parsing trees and inserted a GCN based sub-network to the encoder of sequence-to-sequence machine translation models. \citet{zhang2018graph} applied GCN on pruned syntactic dependency trees for relation extraction. \citet{zhang2019aspect} proposed to use GCN over syntactic dependency trees for aspect-based sentiment classification. \citet{vashishth2019incorporating} applied similar idea to derive word embeddings based on GCN. Furthermore, the tree-LSTM model \cite{tai2015improved} and sent-LSTM \cite{zhang2018sentence} model can also be regarded as implementation of GNN because they both explored recurrent message passing algorithm over tree-structured text. To summarize, the methods in the first group utilize the intrinsic linguistic properties of a sentence to guide the graph building, and then employ GNN to learn better representation of text. 

On the other hand, the second group of studies build graphs in a more heuristic way (e.g. whether an entity appears in a sentence or paragraph), and over wider range of context (e.g. multiple documents). \citet{de2018question} and \citet{xiao2019dfgn} both constructed graphs over entities in documents and capitalized GCN to achieve reasoning over multiple documents. Later, \citet{tu2019hdegraph} proposed to include nodes representing documents to the graph to better model the global information presented in the context. \citet{fang2019hierarchical} built a hierarchical graph consisting of entity nodes, sentence nodes and paragraph nodes for multi-hop reasoning over multiple paragraphs. All these methods attained strong performance on multi-hop reading comprehension tasks. Similar idea was also explored for text classification task \cite{yao2018graph}. The methods in this group aim to learn relational information presented in very long context to achieve reasoning ability by reformatting the context into graph structure.

When dealing with graph nodes represented with sequences (multiple tokens in an entity or a sentence or a paragraph), all previous studies convert the sequence into a feature vector. Then it is possible to apply existing GNN algorithms. However, the GSN proposed in this paper presents a new model to directly conduct message passing algorithm over sequences on graph nodes.

\section{Methodology}
This section starts with a brief introduction on 
GNN. Then, we introduce the proposed GSN and how it is implemented with an emphasis on its difference with existing GNN variants. Finally, we elaborate on how to apply GSN to NLP tasks that require reasoning over sequences.

\subsection{Graph Neural Network}

Assume a graph represented by $\mathcal{G} = (\mathcal{V}, \mathcal{E})$; $\mathcal{V}$ defines a set of N nodes with each node $\mathbf{v}_i$ denoting a $D$-dimensional feature vector; $\mathcal{E}$ defines a set of edges connecting two of the $N$ nodes. Here, we only consider undirected connections between nodes.

GNN is designed for machine learning tasks with structural data that can be represented by a graph to inform the relational information among nodes. 
GNN has two basic operations that can be named as \textit{aggregation} and \textit{combination} in contrast to convolution and pooling in CNN \cite{hamilton2017inductive, xu2018powerful}. One step of these two operations is usually called a hop, and the computation of $k$-th hop can be formulated with \textit{aggregation} and \textit{combination} respectively:
\begin{equation}
    \mathbf{z}_{i}^{k} = f_{agg} \left (\left \{\mathbf{v}_j^{k-1} : j \in \mathcal{N}(i) \right \} \right),
\end{equation}
\begin{equation}
    \mathbf{v}_{i}^{k} = f_{com} \left ( \mathbf{v}_i^{k-1}, \mathbf{z}_i^{k} \right),
\end{equation}
where $f_{agg}$ and $f_{com}$ represent the \textit{aggregation} and \textit{combination} operation respectively. $\mathcal{N}(i)$ is the neighboring nodes of  node $i$. $\mathbf{v}_i^{k}$ is the node representation learned after $k$-th hop. The \textit{aggregation} step collects information from neighboring nodes, while the \textit{combination} step fuses the collected information with the representation of the current node. For example, GCN implements these two steps in one formula:
\begin{equation}
    \mathbf{v}_{i}^{k} = \text{Proj}( \frac{1}{d_i} \sum_{j \in \mathcal{N}(i) \cup i} {\mathbf{v}_{j}^{k-1}}),
\end{equation}
where $\text{Proj}$ is a linear layer with a specific activation function and $d_i$ is the degree of node $i$.

\subsection{Graph Sequential Network}
Assume a different graph represented by $\mathcal{G}_s = (\mathcal{V}, \mathcal{E})$; $\mathcal{V}$ defines a set of N nodes with each node $\mathbf{V}_i$ however denoting a sequence of feature vectors $[\mathbf{v}_i^1, \mathbf{v}_i^2, \cdots, \mathbf{v}_i^{l_i}]$; $l_i$ is the sequence length of node $i$; $\mathbf{v}_i^j$ is a $D$-dimensional feature vector. $\mathcal{E}$ also defines a set of edges connecting two of the $N$ nodes.

Like previous GNN variants, GSN also implements a two-step computation process: \textit{aggregation} and \textit{combination}, which can be formulated by
\begin{equation}
    \mathbf{Z}_{i}^{k} = f_{agg} \left (\left \{\mathbf{V}_j^{k-1} : \forall j \in \mathcal{N}(i) \right \} \right),
\end{equation}
\begin{equation}
    \mathbf{V}_{i}^{k} = f_{com} \left ( \mathbf{V}_{i}^{k-1}, \mathbf{Z}_{i}^{k} \right),
\end{equation}
where $k$ indicates $k$-th computation step. Still, the \textit{aggregation} step calculates structure-aware feature representations $\mathbf{Z}_{i}^{k}$ from the neighborhood of node $i$, and the \textit{combination} step fuses the $\mathbf{Z}_{i}^{k}$ with node $i$'s current feature representation.

To enable aggregation and combination over nodes specified by a sequence of feature vectors, we design new aggregation and combination functions and put them in one formula
\begin{equation}
    \mathbf{V}_{i}^{k} = f_{com} ( f_{coattn}(\mathbf{V}_i^{k-1}, \mathbf{V}_j^{k-1}) ),
    \label{eq:comb}
\end{equation}
where $f_{coattn}$ defines the co-attention based aggregation function. For $f_{com}$, there are two choices: max-pooling $\max_{\forall j \in \mathcal{N}(i) \cup i}$ or average pooling $\frac{1}{d_i} \sum_{j \in \mathcal{N}(i) \cup i}$ ($d_i$ is the degree of node $i$). We can also extend GSN to multi-relational setting as in \cite{schlichtkrull2018modeling}, where there are multiple types of edges. Then, the message passing algorithm with max-pooling based combination becomes
\begin{equation}
    \mathbf{V}_{i}^{k} = \frac{1}{\left |\mathcal{R}\right |}\sum_{r \in \mathcal{R}} \max_{\forall j \in \mathcal{N}^r(i) \cup i} ( f_{coattn}^r(\mathbf{V}_i^{k-1}, \mathbf{V}_j^{k-1}) ),
    \label{eq:multi-rel}
\end{equation}
where $\mathcal{R}$ is the set of all relation types and $\left |\mathcal{R}\right |$ is its size; $\mathcal{N}^r(i)$ is node $i$'s neighbor set, and $f_{coattn}^r$ is the parametrized aggregation function under relation $r$.

There are several ways of implementation for co-attention \cite{xiong2016dynamic, seo2016bidirectional, zhong2019coarse}. Instead of the Recurrent Neural Networks (RNN) based co-attention in \cite{xiong2016dynamic, zhong2019coarse}, we choose the Bidirectional Attention Flow (BiDAF) \cite{seo2016bidirectional} as the co-attention implementation for GSNs for the following reasons: 1) it introduces much less weight parameters compared to RNN (bidirectional) based co-attention. 2) it is much faster than the RNN based co-attention especially when the graph is dense (meaning the graph has almost maximum number of edges that it can have). Our implementation of BiDAF can be summarized in Algorithm \ref{alg:1} (we remove the node and layer indices for clarity). We assume the input and output feature dimensions are both $D$, however it can be adjusted.

For each layer of GSN, the only weight parameters introduced are $\text{Proj}_{i}$ and $\text{Proj}_{o}$; the output size of $\text{Proj}_{i}$ is 1 so it is negligible; the number of parameters of $\text{Proj}_{o}$ is $4D^2$ when input and output feature dimensions are the same.

\RestyleAlgo{boxruled}
\LinesNumbered
\begin{algorithm}[t]
  \caption{Implementation of $f_{coattn}$.\\
  $i$: and :$j$ represent the $i$th row and $j$th column of a matrix respectively; \\
  $\odot$ stands for element-wise multiplication; \\
  ``[;]'' represents vector concatenation; \\
  ``$\text{max}_\text{row}$'' represents taking maximum values over rows of a matrix.}
  \label{alg:1}
  \SetKwInOut{Input}{Input}
  \SetKwInOut{Output}{Output}
  \Input{current node $\mathbf{C} \in \mathbb{R}^{T \times D}$ and one of its neighbor $\mathbf{S} \in \mathbb{R}^{L \times D}$}
  \Output{neighbor-aware representation $\mathbf{O}=f_{coattn}(\mathbf{C}, \mathbf{S}) \in \mathbb{R}^{T \times D}$}
  $\mathbf{M}_{i,j} = \text{Proj}_{i}([\mathbf{S}_{i:};\mathbf{C}_{j:};\mathbf{S}_{i:} \odot \mathbf{C}_{j:}])$, and $\mathbf{M} \in \mathbb{R}^{L \times T} $\;
  $\tilde{\mathbf{S}}_{j:} = \sum_k {\mathbf{a}_{jk} \mathbf{S}_{k:}}$, and $\mathbf{a}_j=\text{softmax}(\mathbf{M}_{:j})$\ and $\tilde{\mathbf{S}} \in \mathbb{R}^{T \times D}$\;
  $\tilde{\mathbf{C}}_{j:} = \sum_k {\mathbf{b}_{k} \mathbf{C}_{k:}}$, and $\mathbf{b}=\text{softmax}(\text{max}_\text{row}(\mathbf{M}))$ and $\tilde{\mathbf{C}} \in \mathbb{R}^{T \times D}$\;
  $\tilde{\mathbf{O}}_{j:} = [\mathbf{C}_{j:}; \tilde{\mathbf{S}}_{j:}; \mathbf{C}_{j:} \odot \tilde{\mathbf{S}}_{j:}; \mathbf{C}_{j:} \odot \tilde{\mathbf{C}}_{j:}]$, and $\tilde{\mathbf{O}} \in  \mathbb{R}^{T \times 4D} $\;
  $\mathbf{O} = \text{Proj}_{o}(\tilde{\mathbf{O}})$, and $\mathbf{O} \in  \mathbb{R}^{T \times D} $\;
\end{algorithm}

\subsection{Applications on NLP tasks}

Some NLP tasks require reasoning over multiple sentences/paragraphs, such as 
multi-hop machine reading comprehension or fact verification over multiple sentences/documents \cite{yang2018hotpotqa, thorne2018fever}. 
Previous studies have shown that by applying GNN to the graph with sequence (phrase, sentence or document) embeddings as nodes can improve the performance of these tasks \cite{xiao2019dfgn, tu2019select, zhou2019gear, fang2019hierarchical}. Instead of summarizing sequences into vectors and using them for graph node initialization, our proposed GSN avoids the sequence summarization and directly take sequence features as graph node representation. The co-attention based message passing of GSN can learn neighbor-aware representations of the current node. The current node acts as the context sequence and each of its neighbor acts as the query sequence as in co-attention for MRC  \cite{seo2016bidirectional}. Thus the GSN enables aggregation of relational information among sequences and strengthens the model's reasoning ability over sequences. Furthermore, based on its sequential output, GSN also makes sequential labeling tasks possible, such as start and end positions prediction for extraction based QA tasks, while it is impossible for current GNNs variants to achieve this. This property could bring more potential to sequential labeling tasks in NLP which requires complex reasoning. 

\section{Experiments and Results}
In this section we validate the efficacy of our proposed GSN on three NLP tasks: multi-hop MRC span extraction, multi-hop MRC supporting sentence prediction and fact verification using two data sets: HotpotQA \cite{yang2018hotpotqa} and FEVER \cite{thorne2018fever}. Our goal in this study is not to achieve the state-of-the-art performance on these two data sets, but rather to show the effectiveness of the proposed GSN when compared to existing GNN models.

\subsection{HotpotQA data set}
HotpotQA is the first multi-hop QA data set taking the explanation ability of models into account. HotpotQA is constructed in the way that crowd workers are presented with multiple documents and are asked to provide a question, corresponding answer and supporting sentences used to reach the answer. There are about 90K training samples, 7.4K development and test samples.
HotpotQA presents two tasks: answer span prediction (to extract a text span from the context) and supporting facts prediction (to predict whether a sentence supports the answer or not). Models are evaluated based on Exact Match (EM) and $F_1$ score of the two tasks. Joint EM and $F_1$ scores are used as the overall performance measurements, which encourage the model to be accurate on both tasks for each example. In this study we apply the proposed GSN to the distractor setting of the data set.

Since each HotpotQA example comes with 10 documents with 8 of them are distraction, and only the remaining 2 are useful for answering the question, we choose to only use the 2 gold documents as the context of each question to focus on comparing GNN and GSN. 
The 2 gold documents have multiple sentences: some of them are annotated as supporting sentences and the answer span resides in one of the sentence. We concatenate the 2 gold documents as in \cite{xiao2019dfgn, tu2019select} and use BERT (base uncased) \cite{devlin2018bert} to encode the ``[CLS]+question+[SEP]+context+[SEP]'' input. A sentence extractor is applied on the output of BERT to get the sequential output of each sentence with pre-calculated sentence start and end indices. 

To build a graph on these sentences, we extracted named entities (NE) and noun phrases (NP) and the question, and two sentences are connected if 1) they come from the same document; 2) they come from the different documents but share the same NEs or NPs; 3) they come from the different documents but both have one or more NEs or NPs appeared in the question. We treat those three types of edges differently as in Equation \ref{eq:multi-rel}.
Then we apply the proposed GSN on the built graph and get the updated sequential representation of each node. Finally, all sentences can be re-concatenated for the the span prediction task by predicting the start and end indices, or summarized into fixed-dimensional vectors for supporting sentence prediction. The former task is optimized with a cross entropy (CE) loss while the later with a binary CE loss. We also jointly optimize the two tasks together by weighted summation of the two loss items. Since the point of our paper is to show the efficacy of the proposed GSN model,
we only show results on development set; getting numbers on test set requires several other modules \cite{xiao2019dfgn, tu2019select} that are unrelated to our proposed GSN.

\subsection{Results on HotpotQA}

\subsubsection{Experimental Settings}
We present the results on HotpotQA data set in three experimental settings as we want to show the proposed GSN performs better in different tasks compared to baseline models. In the first setting, 
we use multi-relational GSN model on top of the BERT sequential output to only predict supporting sentences from the context. 
We compare it with the model based on multi-relational GCN \cite{schlichtkrull2018modeling} over early summarized feature vectors to learn structure-aware sentence representation, which has been employed in previous studies for multi-hop MRC \cite{de2018question, xiao2019dfgn, tu2019select}. 

In the second setting, multi-relational GSN model is applied on top of BERT sequential output to only predict answer span, which is a sequential labeling task on the output of the GSN model. Note that standard GNN models are incompetent at this task because the sentences are summarized into vectors and there is no way to make predictions on position (token) level with GNN output. Thus, the baseline model we compare with is to directly classify the tokens in BERT sequential output.

In the third setting, we train the GSN model by jointly predicting answer span at token level and supporting sentences at sentence level. We compare with the baseline model which jointly trains answer span prediction and supporting sentence prediction baseline models.

All results are the average of five runs with different random seeds, and we also show the standard deviation of the numbers. Please refer to supplementary materials for details about implementation and hyperparameter settings.

\subsubsection{Results}
\label{sec:result}

We report the results using the best hyperparameters for each experimental setting. First, in Table~\ref{tab:single} we show the results of the baseline GCN-based model and the proposed GSN-based model for answer prediction only and supporting sentence prediction only tasks in terms of EM and $F_1$ score. The results show that the proposed GSN-based models perform better on both tasks with strong statistical significance compared to the baseline GCN-based model. The improvement on EM score is slightly more significant, indicating GSN-based models are better at finding complete answer span or supporting sentences than the baseline models.

Table~\ref{tab:joint} demonstrates the results when we make predictions of both answer span and supporting sentences over the GSN output. Compared to the baseline model, more improvement is attained than models trained only on one task, especially for the supporting sentence prediction task. With joint training, the performance on answer span prediction drops while the performance on supporting sentence prediction increases. Actually we observed better joint EM and $F_1$ scores with joint training compared to separate training for both baseline models and GSN-based models. Thus, joint training still boosts the performance overall, because the joint trained models find the correct answer and supporting sentences of a question simultaneously.

\begin{table}[]
\centering
\caption{Results comparison with average and standard deviation of five runs. ``ANS-only'' and ``SUP-only'' indicate the model is only trained on two separate tasks.}
\resizebox{\columnwidth}{!}{%
\begin{tabular}{|c|c|c|c|c|}
\hline
         & \multicolumn{2}{c|}{ANS-only} & \multicolumn{2}{c|}{SUP-only} \\ \hline
         & EM          & $F_1$      & EM          & $F_1$      \\ \hline
baseline & 63.87$\pm$0.16      & 77.69$\pm$0.16     & 62.14$\pm$0.16      & 88.94$\pm$0.07      \\ \hline
GSN      & 64.39$\pm$0.06       & 78.27$\pm$0.10      & 62.96$\pm$0.14       & 89.29$\pm$0.07      \\ \hline
\end{tabular}}
\label{tab:single}
\end{table}

\begin{table*}[]
\centering
\caption{Results comparison with average and standard deviation of five runs. ``ANS'', ``SUP'' and ``JOINT'' indicate the jointly trained models' performance in terms of measurements on answer span prediction, supporting sentence prediction and joint tasks.}
\resizebox{0.8\textwidth}{!}{%
\begin{tabular}{|c|c|c|c|c|c|c|}
\hline
         & \multicolumn{2}{c|}{ANS} & \multicolumn{2}{c|}{SUP} & \multicolumn{2}{c|}{JOINT} \\ \hline
         & EM          & $F_1$      & EM          & $F_1$      & EM           & $F_1$       \\ \hline
baseline & 62.99$\pm$0.16       & 76.90$\pm$0.31      & 61.35$\pm$0.17       & 88.73$\pm$0.09      & 41.76$\pm$0.40        & 69.64$\pm$0.28       \\ \hline
GSN      & 63.56$\pm$0.31       & 77.26$\pm$0.11      & 63.26$\pm$0.16       & 89.35$\pm$0.04      & 43.51$\pm$0.27        & 70.43$\pm$0.14       \\ \hline
\end{tabular}
}
\label{tab:joint}
\end{table*}

\subsubsection{Analysis}

\begin{table}[]
\centering
\caption{Effect of the number of GSN layers on the performance.}
\resizebox{\columnwidth}{!}{%
\begin{tabular}{|c|c|c|c|c|c|c|}
\hline
         & \multicolumn{2}{c|}{ANS-only} & \multicolumn{2}{c|}{SUP-only} & \multicolumn{2}{c|}{JOINT} \\ \hline
         & EM          & $F_1$      & EM          & $F_1$      & EM           & $F_1$       \\ \hline
1 layer & \textbf{64.38}       & \textbf{78.24}      & 62.36       & 89.09      & 42.70        & 70.23       \\ \hline
2 layer      & 64.29       & 78.06      & 62.38       & 89.21      & \textbf{43.81}        & \textbf{70.56}       \\ \hline
3 layer      & 64.23       & 77.96      & \textbf{63.07}       & \textbf{89.37}      & 43.36        & 70.24       \\ \hline
\end{tabular}
}
\label{tab:layer}
\end{table}

\begin{table}[]
\centering
\caption{Results comparison between mean-pooling and max-pooling combination functions.}
\resizebox{\columnwidth}{!}{%
\begin{tabular}{|c|c|c|c|c|c|c|}
\hline
         & \multicolumn{2}{c|}{ANS} & \multicolumn{2}{c|}{SUP} & \multicolumn{2}{c|}{JOINT} \\ \hline
         & EM          & $F_1$      & EM          & $F_1$      & EM           & $F_1$       \\ \hline
mean-pooling & 63.47       & 77.04      & 61.93       & 89.03      & 42.61        & 70.01       \\ \hline
max-pooling      & 63.92       & 77.39      & 63.36       & 89.39      & 43.81        & 70.56       \\ \hline
\end{tabular}
}
\label{tab:comb}
\end{table}

\textbf{Effect of number of layers:} we investigated the influence of the number of layers (hops) of the proposed GSN on the performance. We changed the number of layers from 1 to 3, and record the same set of measurements with all three experimental settings. The results of the best random seed are presented in Table~\ref{tab:layer}. We only show the joint measurements for the joint training experiment. 

The results show that the three tasks with different training objectives demonstrate totally different performance patterns in terms  of the number of GSN layers: the answer prediction task achieves the best result with 1-layer GSN (still better than without GSN as shown in Table~\ref{tab:single}), while the supporting sentence prediction task requires a 3-layer GSN. When jointly training both tasks, the 2-layer GSN gives the best performance. This pattern is reasonable because the powerful BERT based encoder possibly learns good contextual representation on token level. This learned representation can benefit the answer prediction task which is also on token level, thus less graph based reasoning is required. Similarly, some recent studies \cite{min2019compositional} found that considerable amount of questions in HotpotQA can be answered without multiple hops. However, it is not the case for supporting sentence prediction as it requires the model to find sentences that could be far from each other in the context. On the contrast, our proposed GSN is suitable to model the relational information among sentences, thus we observe that more layers give better results for supporting sentence prediction.

\textbf{Effect of combination function:} we have introduced two choices of combination function in Equation~\ref{eq:comb}. Actually the results in section \ref{sec:result} are all achieved with the max-pooling based combination function as we found it is always better than the mean-pooling alternative. To demonstrate this, we show the results comparison only for the last experiment settings: the joint training strategy with the best random seeds. Table~\ref{tab:comb} gives the detail of the comparison.

\textbf{Indication:} as discussed previously, the two tasks on the HotpotQA data set can be regarded as a sequential labeling task (answer prediction) and a node classification (supporting sentence prediction) respectively. Through experiments with different settings, we have shown our proposed GSN model can 1) deal with sequential labeling tasks which require reasoning over context; existing GNN based models are unable to tackle such tasks. 2) attains better performance than GNN based models on node classification tasks with sequences as nodes.

\subsection{FEVER data set}
The FEVER data set is provided by the FEVER shared task 1.0\footnote{http://fever.ai/2018/task.html}. The goal of FEVER shared task is to develop automatic methods to extract evidence from Wikipedia and verify human-generated claims given these evidence. In this study, we focus on the later task: fact verification. We used the same evidence extraction output from the baseline system \cite{zhou2019gear}. The resulting data set has a claim and multiple sentences for each sample, and the model needs to predict whether the evidence \textit{support} or \textit{refute} the claim or there is \textit{not enough information} to make the prediction. In total, there are about 145K samples in training set, 20K samples in development and test set respectively. The baseline system \cite{zhou2019gear} employed BERT to encode each claim-evidence pair and proposed a GAT based evidence aggregation model to exploit the relational information among multiple pieces of evidence. Then, a graph is built to connect nodes representing the encoded embedding of every claim-evidence pair. Finally, the fact verification becomes a graph classification task over the graph. An attention based read-out layer is designed to obtain a graph feature vector which is sent to a classifier to predict the target.

In our experiments, we follow exactly the same data preprocessing scripts with the baseline system \footnote{https://github.com/thunlp/GEAR}. Our model design is different from the baseline system in the following aspects: 1) the baseline system trained the BERT-based sentence encoder and GAT model separately while we trained the two parts together. 2) For joint training, to save GPU memory usage, we concatenate all evidence sentences as the context, which is then paired with the claim and sent to BERT. We employ an attention based pooling strategy to get the sentence embedding from the BERT output given the start and end positions of each sentence in the context. We show later in the results that those two modifications give slightly better results than the baseline system. 3) The GAT based evidence aggregator is replaced with our proposed GSN model. To use GSN, there is no need to do the attention based pooling over BERT output to get sentence embedding; instead we directly input the sequential outputs of all sentences to the GSN model. After evidence aggregation, a two-step graph embedding extraction is applied: first step is to convert GSN's output for each sentence to a vector, and second step is to convert all sentences' embeddings to a single vector to predict targets. We use attentive pooling for both steps. For all models, the training objective is CE loss. Please refer to supplementary materials for details about hyperparameter settings.

\subsection{Results on FEVER data set}

For FEVER data set, the test set is blind and the prediction needs to be submitted to codalab \footnote{https://competitions.codalab.org/competitions/18814} for evaluation. The measurements are label accuracy (ACC) to measure the accuracy of fact verification and official FEVER score to measure the label accuracy conditioned on providing at least one complete set of evidence. We report our best results on the development set and their corresponding numbers on the test set, and compare our proposed GSN-based model with the reported numbers in the baseline paper \cite{zhou2019gear} and our implementation of the baseline system.

Table~\ref{tab:fever} shows the results of three systems on both the FEVER development and test set. Our re-implementation of the baseline system gets slightly better numbers than those reported by \citet{zhou2019gear}. And our proposed GSN based system improves over the baseline system by more than 1\% in terms of both ACC and FEVER score. Since the fact verification task can be regarded as a graph classification task, we further prove the proposed GSN is able to achieve better performance than GNN based models when using sequences on graph nodes.

\begin{table}[]
\centering
\caption{Results on FEVER development and test sets.}
\resizebox{\columnwidth}{!}{%
\begin{tabular}{|c|c|c|c|c|}
\hline
         & \multicolumn{2}{c|}{dev} & \multicolumn{2}{c|}{test} \\ \hline
         & ACC          & FEVER      & ACC         & FEVER      \\ \hline
\makecell{baseline \\ \cite{zhou2019gear}} & 73.67      & 68.69     & 71.01      & 65.64      \\ \hline
baseline (ours) & 73.72      & 69.26     & 70.80      & 65.88      \\ \hline
GSN      & \textbf{74.89}       & \textbf{70.51}      & \textbf{72.00}       & \textbf{67.13}      \\ \hline
\end{tabular}}
\label{tab:fever}
\end{table}

\subsection{Visualization of attention in GSNs}

\begin{figure}[t]
    \centering
    \includegraphics[width=1.0\linewidth]{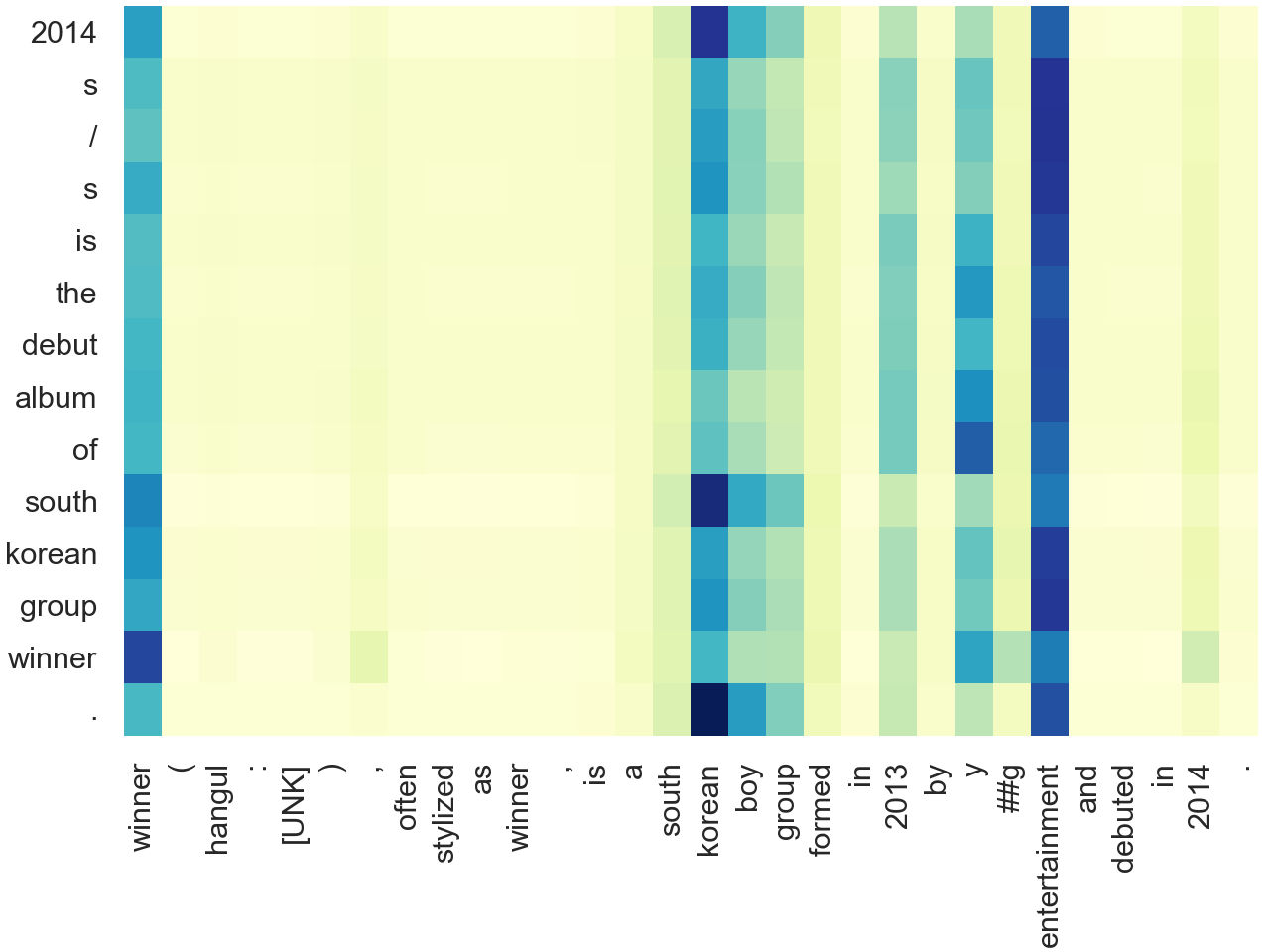}
    \caption{Heatmap of an attention matrix in the BiDAF based co-attention between two sentences. The ``[UNK]'' token is caused by Hangul characters.}
    \label{fig:attn}
\end{figure}

To illustrate how the co-attention based message passing algorithm works, in Figure~\ref{fig:attn} we show the heatmap of an attention matrix (matrix $M$ in Algorithm 1) between two nodes on the graph built from a sample in the HotpotQA development set. The question of this sample is ``2014 S/S is the debut album of a South Korean boy group that was formed by who?''. The current node is ``2014 S/S is the debut album of South Korean group WINNER.'' (after tokenization as shown in the y-axis of Figure~\ref{fig:attn}), and its neighbor is ``Winner (Hangul: 위너), often stylized as WINNER, is a South Korean boy group formed in 2013 by YG Entertainment and debuted in 2014.'' (after tokenization as shown in the x-axis of Figure~\ref{fig:attn}). The two sentences are from different documents. We apply softmax to the sequence direction of the neighbor node to see the attention pattern of each token in the current node over the tokens in the neighbor node. It is clear that almost all tokens in the current node assign high attention weight to the tokens ``y'' and ``entertainment'' because they are the start and end positions of the answer span. Meanwhile, the ``winner'' and ``south korean boy group'' are also attended by tokens in the current node because they act as the bridging entities leading to the final answer ``YG Entertainment''. This figure clearly shows our GSN-based models can find multiple pieces of useful information with message passing over sequences.
We include more visualization in supplemental materials.

\section{Conclusion}

This paper proposes graph sequential network as a novel neural architecture to facilitate reasoning over graphs with sequential data on the nodes. We develop a new message passing algorithm based on co-attention between two sequences on graph nodes. The scheme avoids the information loss inherent in the pooling based early summarization of existing GNN-based models, and improve the reasoning ability on sentence level. Through experiments on HotpotQA and FEVER, both of which require the model to perform multi-hop reasoning, we show that our proposed GSN attains better performance than existing GNNs on different types of tasks. For future work we would like to apply GSN to other applications in NLP that require complex reasoning.

\bibliography{acl2020}
\bibliographystyle{acl_natbib}

\end{document}